%% file: main.tex
\documentclass[runningheads]{llncs}
\usepackage{graphicx}
\usepackage{amsmath,amssymb} 
\usepackage{color}
\usepackage[width=122mm,left=12mm,paperwidth=146mm,height=193mm,top=12mm,paperheight=217mm]{geometry}
\newcommand{\vs}{$\rightarrow$}

\begin{document}
\pagestyle{headings}
\mainmatter
\def\ECCV16SubNumber{}  

\title{Learning the Roots of Visual Domain Shift} 

\authorrunning{Tatiana  Tommasi, Martina Lanzi, Paolo  Russo, Barbara  Caputo}

\author{Tatiana  Tommasi$^1$, Martina Lanzi$^2$, Paolo  Russo$^2$, Barbara  Caputo$^2$}
\institute{$^1$ University of North Carolina at Chapel Hill, NC, USA\\ $^2$ University of Rome La Sapienza, Dept. of Computer, Control and Management Engineering, Rome, Italy}

\maketitle

\begin{abstract}
\input{abstract}

\end{abstract}

\section{Introduction}
\label{sec:intro}
\input{intro}\vspace{-3mm}

\section{Related Work}\vspace{-2mm}
\label{sec:related}
\input{related}\vspace{-5mm}

\section{Domainness Prediction}\vspace{-3mm}
\label{sec:strategy}
\input{strategy}\vspace{-2mm}

\section{Domainness Analysis}\vspace{-5mm}
\label{sec:analysis}
\input{analysis}

\section{Exploiting Domainness Levels for Adaptation}
\label{sec:experiments}
\input{experiments}

\section{Conclusion}
\label{sec:conclusion}
\input{conclusion}


\bibliographystyle{splncs}
\bibliography{egbib}
\end{document}

%% file: abstract.tex
In this paper we focus on the spatial nature of visual domain shift, attempting to learn \emph{where} 
domain adaptation originates in each given image of the source and target set. 
We borrow concepts and techniques from the CNN visualization literature, and learn 
\emph{domainnes maps} able to localize the degree of domain specificity in images. 
We derive from these maps features related to different domainnes levels, 
and we show that by considering them as a preprocessing step for a domain adaptation 
algorithm, the final classification performance is strongly improved.
Combined with the whole image representation, these features provide 
state of the art results on the Office dataset.

%% file: intro.tex
In 2010 Saenko et al.  imported the notion of domain adaptation from natural language processing to 
visual recognition \cite{Saenko:2010}. They showed how training visual classifiers on data acquired in a 
given setting, and testing them in different scenarios, leads to poor performance because the training 
and test data belong to different visual domains.
Since then, domain adaptation has become a widely researched topic. The vastly dominant trend is to 
summarize images into global features (being them handcrafted BoWs or the most modern CNN-activation values) 
and remove the domain shift through an optimization problem over feature data points distributions.  
This strategy is theoretically sound and effective, as it has been largely demonstrated over the years. 
To give a quantitative estimate of the progress in the field, one might look at the accuracy values obtained 
over the Office-31 dataset, a data collection presented in \cite{Saenko:2010} and quickly become the 
domain adaptation reference benchmark: performance has increased on average from 
27.8\%  \cite{GongSSG12} to 72.9\% in only three years \cite{long_icml_2015}.
While such progress is certainly impressive, it is not fully clear that it is coupled with an equally 
deepened knowledge of the roots of domain shift.

We believe the time is ripe for gaining a better understanding of how visual concepts such as 
illumination conditions, image resolution or background give rise to the domain shift. As these visual 
concepts often have a spatial connotation -- more or less illuminated parts of images, informative object 
parts that are more or less visible, etc. -- our goal is to localize the domain shift 
in the source and target data, or at least to spatially ground it. Is it doable? and if yes, what do we 
learn from it? Could it be used to improve the effectiveness of \emph{any}  domain adaptation algorithm?

\begin{figure}[t]
\begin{center}
\includegraphics[width=0.9\linewidth]{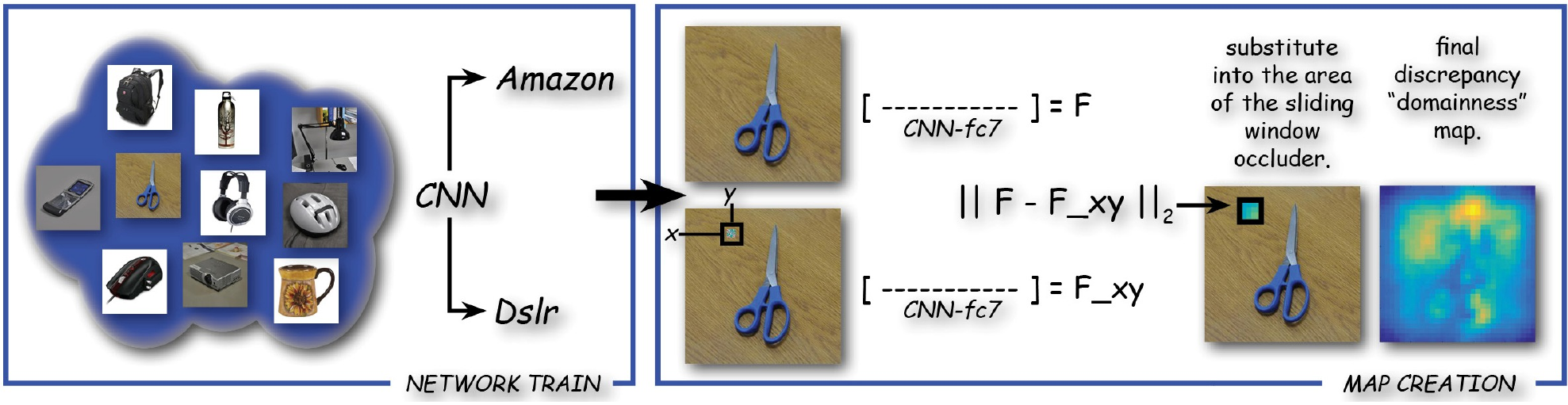} 
\end{center}\vspace{-5mm}
\caption{Domainness maps generation procedure. A CNN network is 
initially trained for domain classification. The obtained model is used as feature extractor both for the 
original and the occluded images. The difference among the obtained representation is saved and visualized 
in the image area, creating the discrepancy map. Yellow/Blue indicate areas at high/low domainness. Best viewed in color.}
\label{fig:overview}\vspace{-5mm}
\end{figure}

This paper attempts to answer these questions.
We first show that by learning to classify visual domains (binary classification on source-target domain pairs), 
it is possible to obtain domain localization maps as a byproduct, where high/low map values indicate high/low 
domain specificity (Figure \ref{fig:overview}, section \ref{sec:strategy}). We dub the score used to define the map \emph{domainness}. 
By analyzing the domainnes map we are able to evaluate the correlation between domain-specificity and object-specificity. 
Depending on the domain-pairs we can identify when the domain shift come mostly from the background and when 
instead it involves the objects (section \ref{sec:analysis}).
Armed with this knowledge,  
we create 3 different features from each image: a low-domainness feature,  a mid-domainness feature and a high-domainness
feature (Figure \ref{fig:domainnesslevels}, section \ref{sec:experiments}). With this strategy each domain-pair becomes a set of 9 pairs. 
We show that by applying domain adaptation over each pair and then recombining the results 
through high level integration, we systematically achieve a substantial increase in performance as opposed to 
previously reported results obtained by the same methods on the whole images. This approach enables us to obtain the 
new state of the art on the Office-31 dataset for unsupervised domain adaptation. 

%% file: related.tex
\emph{Domain Adaptation.} 
The goal of domain adaptation is to compensate the variation among two data distributions, allowing to reuse 
information acquired from a source domain on a new, different but related, target domain. 
Some techniques perform this by simply re-weighting or selecting the samples in the source domain \cite{landmarks,Lim_nips11}, 
or clustering them to search for 
visually coherent sub-domains \cite{Hoffman_ECCV2012,reshape}. Other approaches modify existing source 
classifiers to make them suitable for the target task \cite{DuanICML2009,DuanCVPR2009}, or search for 
transformations that map the source distribution into the target one \cite{Gretton_MMD_2012,Fernando2013b}. 
Different strategies propose to learn both a classification model and a feature transformation 
jointly. Few of them rely on SVM-like risk minimization objectives and shallow representation models 
\cite{Hoffman_ICLR2013,Fernando_Tommasi_Tuytelaars_PRL2015}, while more recent approaches 
leverage over deep learning \cite{Tzeng_ICCV_2015,Tzeng_arxiv_2015,long_icml_2015}. 

Despite their specific differences, all these methods consider the whole image as a data unit, corresponding 
to a sample drawn from a given domain distribution. Some work has been recently
done for dealing directly with image patches within the NBNN framework 
\cite{Tommasi_2013_ICCV,Kuzborskij_CVPR_2016} with promising results. Here we push research further 
in the direction of relating domain adaptation and spatial localities in images: we study how the domain information is 
distributed inside each image, and how to deal with domain-specific and domain-generic image patches.  

\vspace{1mm} \noindent
\emph{CNN Visual Analysis.} A number of works have focused on understanding the representation learned by CNNs. 
A visualization technique which reveals the input stimuli that excite individual feature maps at any layer in 
the model was introduced in \cite{Zeiler_ECCV_2014}. Girshick et al. \cite{girshick2014rcnn} showed 
visualizations that identify which patches within a dataset
are the most responsible for strong activations at higher layers in the model.  Simonyan et al. 
\cite{Simonyan_ICLR_2014} demonstrated how saliency maps can be obtained from CNN by projecting back 
from the fully connected layers of the network. Zhou et al. \cite{Zhou_ICLR_2015} focused on scene 
images, and by visualizing the representation learned by each unit of the network, they showed that 
object detectors are implicitly learned. 

Inspired by these works we introduce an image mask-out procedure to visualize what a domain classification 
network learns and how the domain information is spatially distributed. We are not aware of previous work 
attempting to learn what part of images are more or less responsible for the domain shift.

%% file: strategy.tex
In this section we describe the details of the data-driven visualization approach defined to localize the domain shift in images. 

Given the images of a source/target domain pair we resize them to $256\times256$, and we randomly split them into a 
training and test set. On the training set we learn a CNN for domain recognition: specifically we initialize the parameters of 
conv1-fc7 using the released CaffeNet \cite{jia2014caffe} weights 
and we then further fine-tuned the network for binary classification on the domain labels. 
The test set is extended by replicating
each image many times with small random occluders at different locations: we use $16\times16$ image patches 
positioned on a dense grid with stride 8. This results in about 1000 occluded images per original image. 
Finally we feed both the original and the occluded test images into the defined network and we record the 
difference between their respective fc7 activation values (4096-dimensional output of the seventh fully connected layer after ReLu). 
The $L_2$-norm of this difference is spatially saved in the image inside the occluder area and 
the obtained value for overlapping occluders is averaged defining a smooth discrepancy map with values
rescaled in $\{0,1\}$ (see Figure \ref{fig:overview}). 
We call it \emph{domainness} map: an area of high domainness 
corresponds to a region that highly influences the final domain choice for the image, and thus it 
can be considered as domain-specific. On the other hand, an area of low domainness appears to be 
less relevant for domain recognition, hence more domain-generic. Note that the procedure we propose is 
unsupervised with respect to the object classes depicted in the source and target images. 

%% file: analysis.tex
\begin{figure}[t]
\hspace{-3mm}
\tiny
\begin{tabular}{|cccc|cccc|cccc|}
\hline
\multicolumn{4}{|c|}{A-D} & \multicolumn{4}{|c|}{A-W}& \multicolumn{4}{|c|}{D-W}\\
\hline
A & \includegraphics[width=0.09\linewidth]{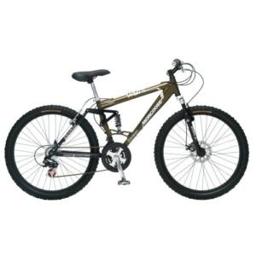} 
& \includegraphics[width=0.115\linewidth]{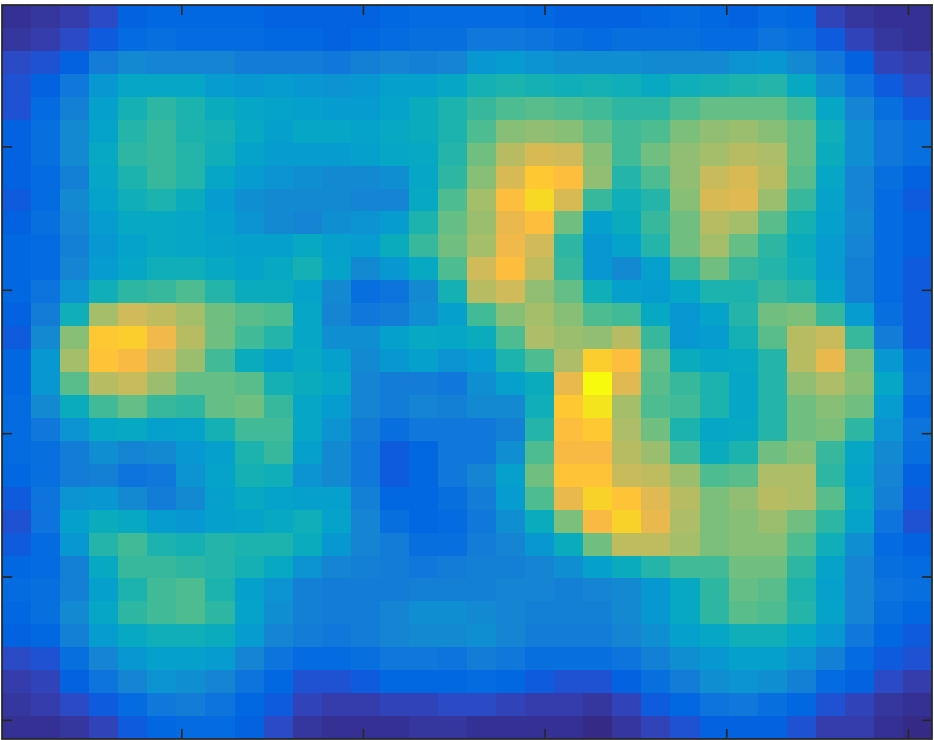} 
& \includegraphics[width=0.09\linewidth]{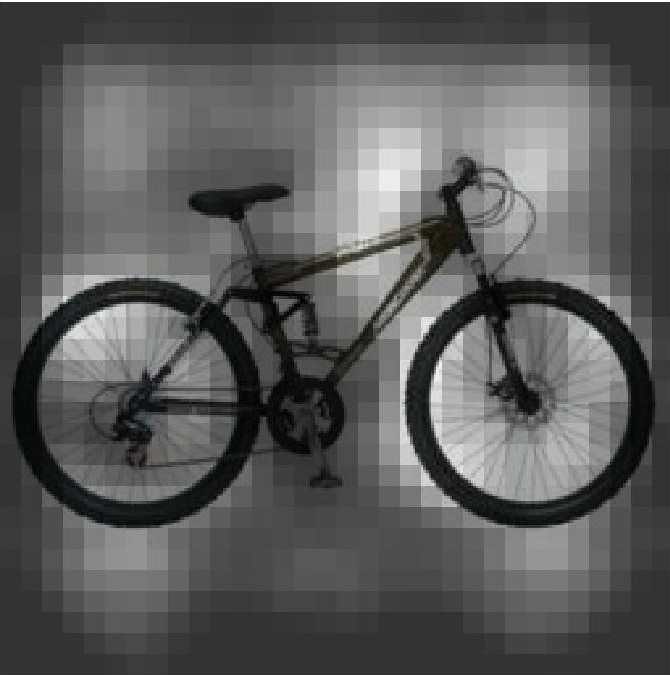} &
A & \includegraphics[width=0.09\linewidth]{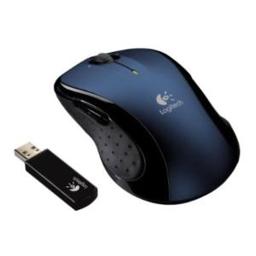} 
& \includegraphics[width=0.115\linewidth]{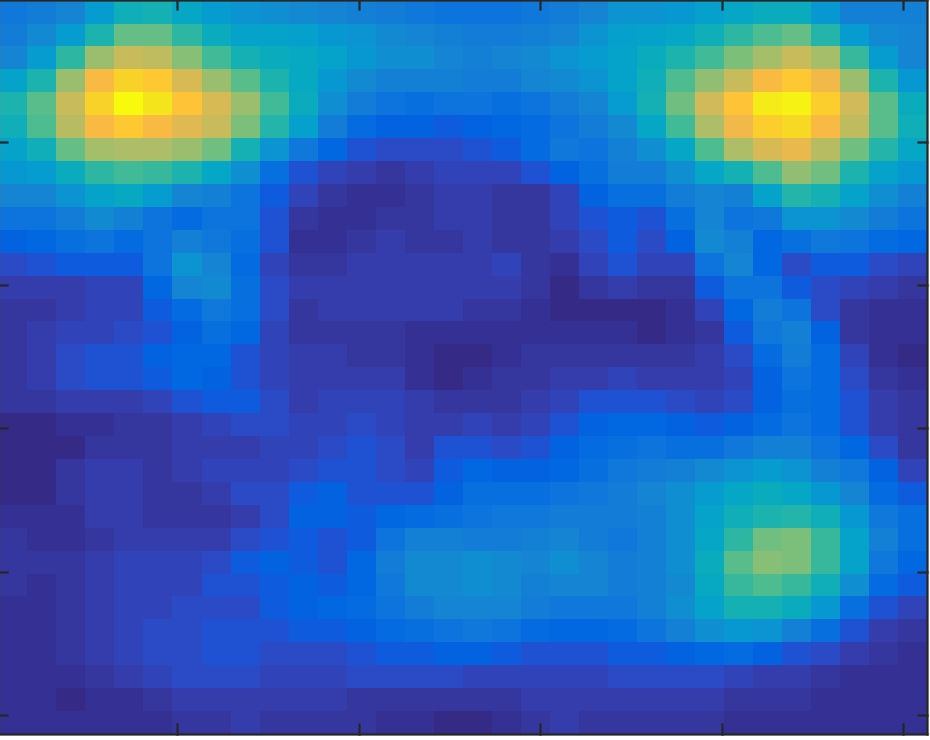} 
& \includegraphics[width=0.09\linewidth]{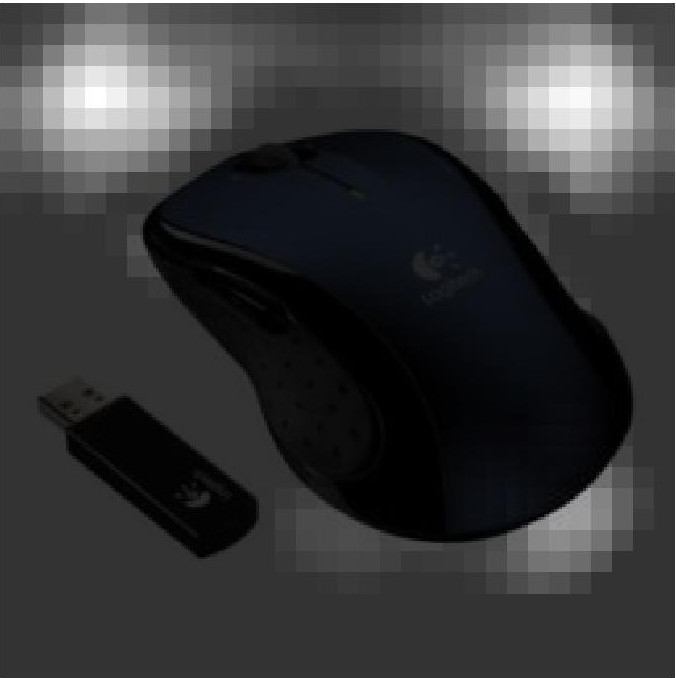} &
D & \includegraphics[width=0.09\linewidth]{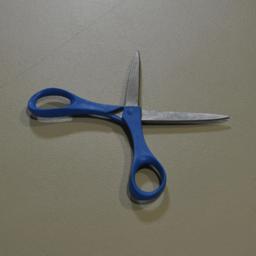} 
& \includegraphics[width=0.115\linewidth]{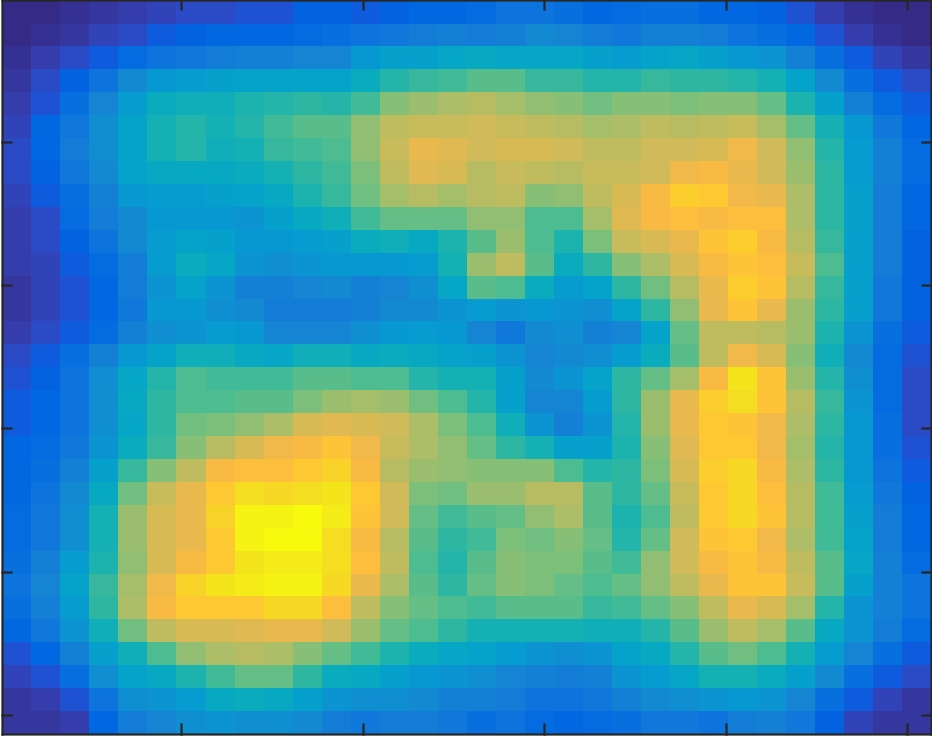} 
& \includegraphics[width=0.09\linewidth]{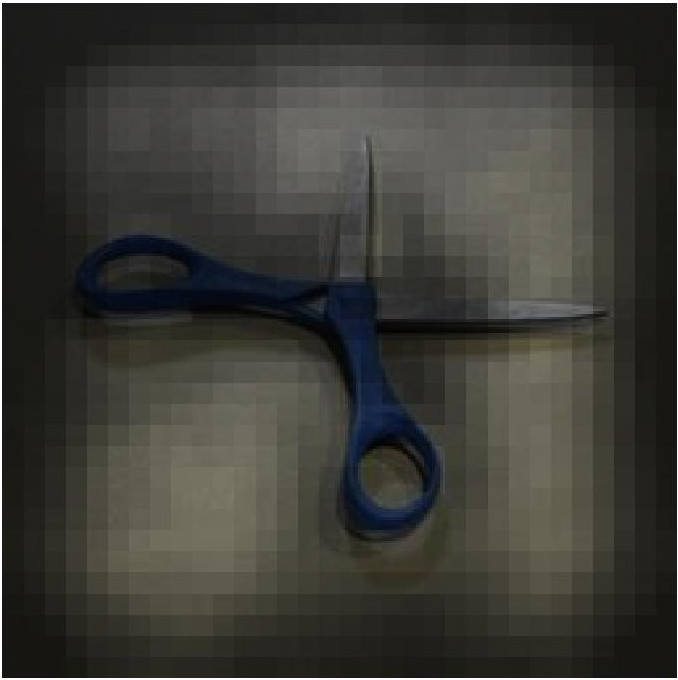} \\ \hline
D & \includegraphics[width=0.09\linewidth]{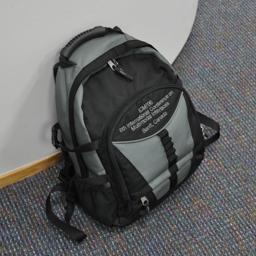} 
& \includegraphics[width=0.115\linewidth]{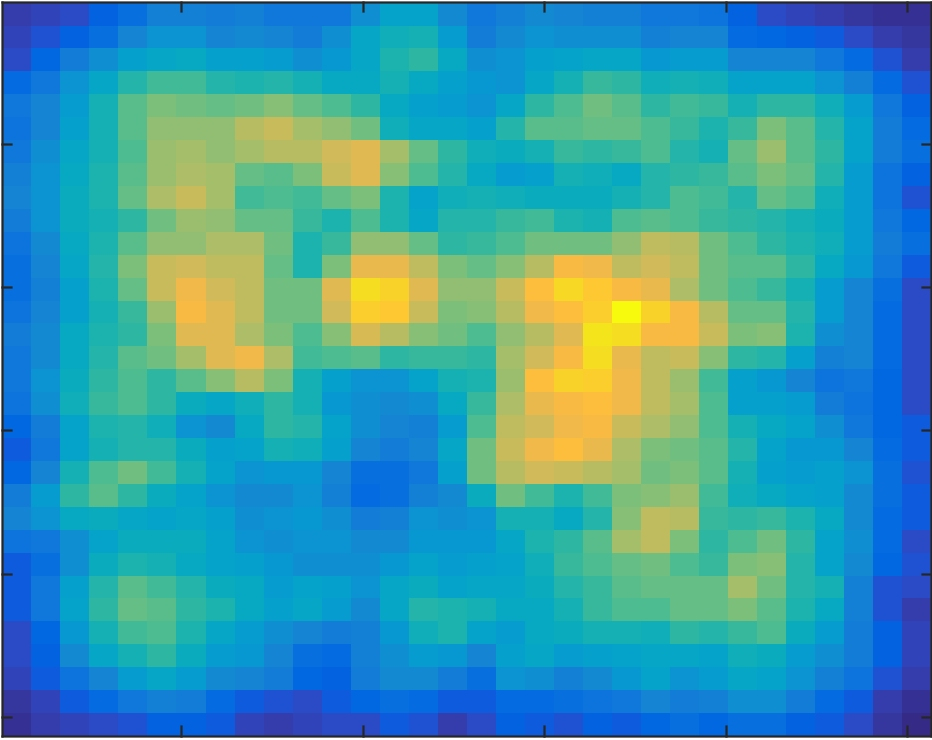} 
& \includegraphics[width=0.09\linewidth]{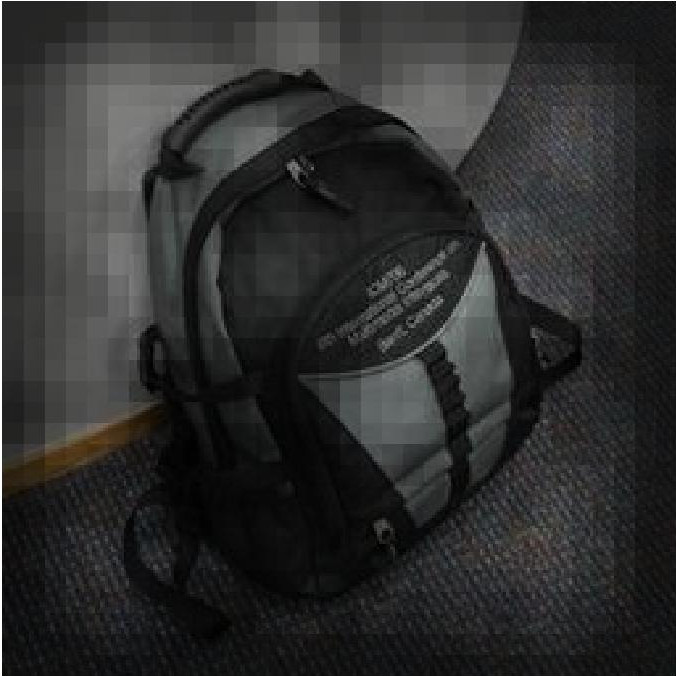} &
W & \includegraphics[width=0.09\linewidth]{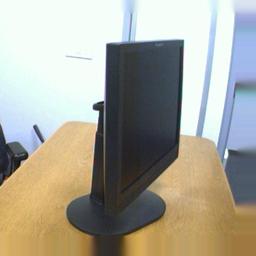} 
& \includegraphics[width=0.115\linewidth]{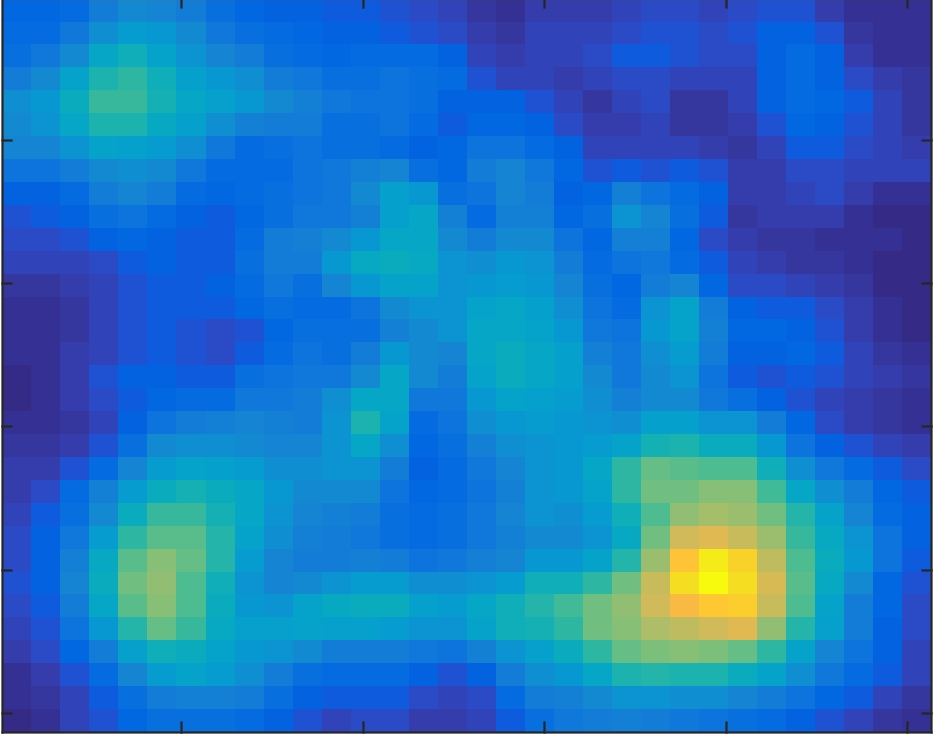} 
& \includegraphics[width=0.09\linewidth]{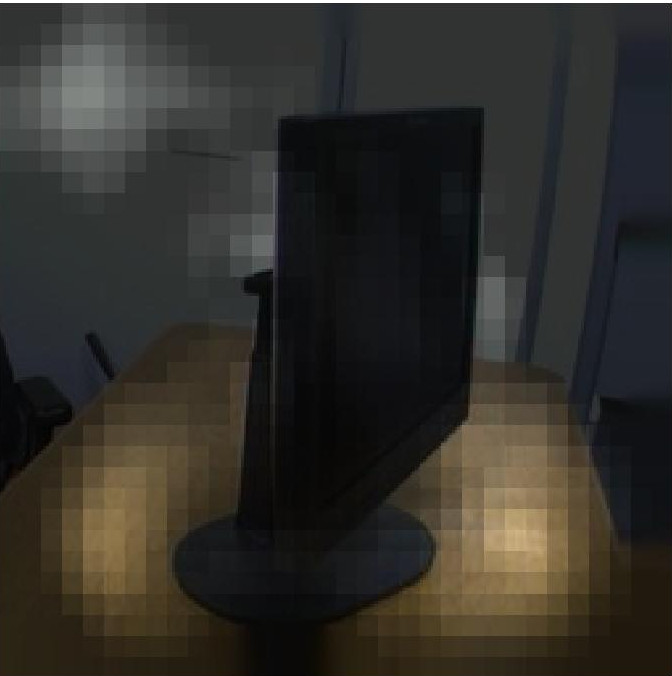} &
W & \includegraphics[width=0.09\linewidth]{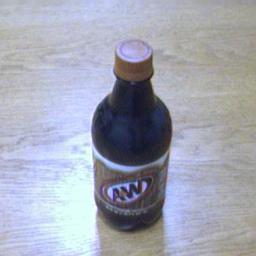} 
& \includegraphics[width=0.115\linewidth]{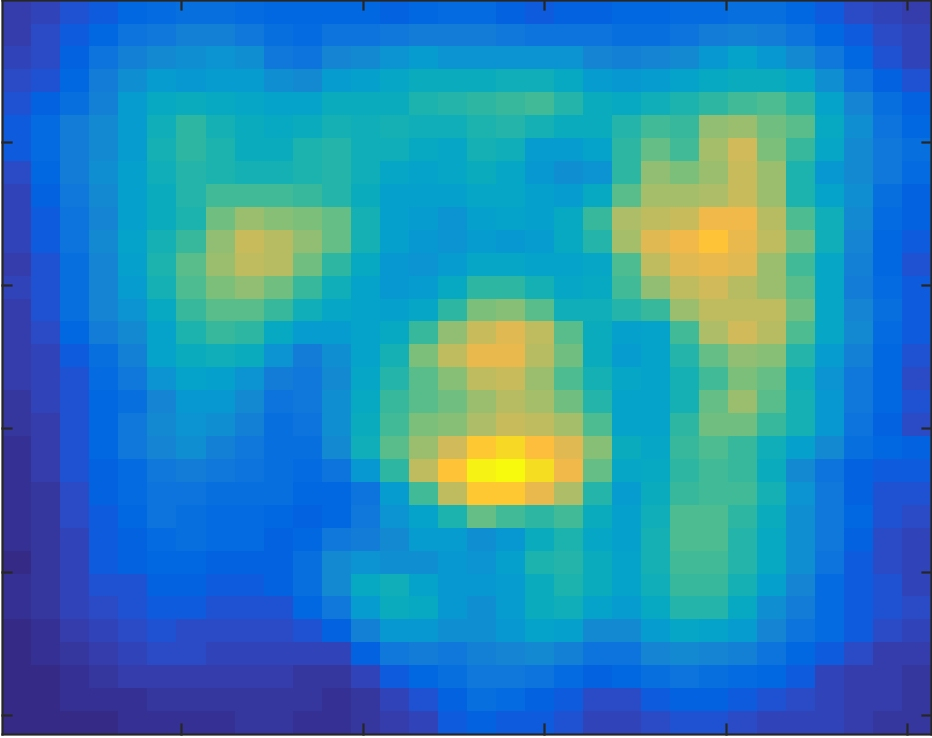} 
& \includegraphics[width=0.09\linewidth]{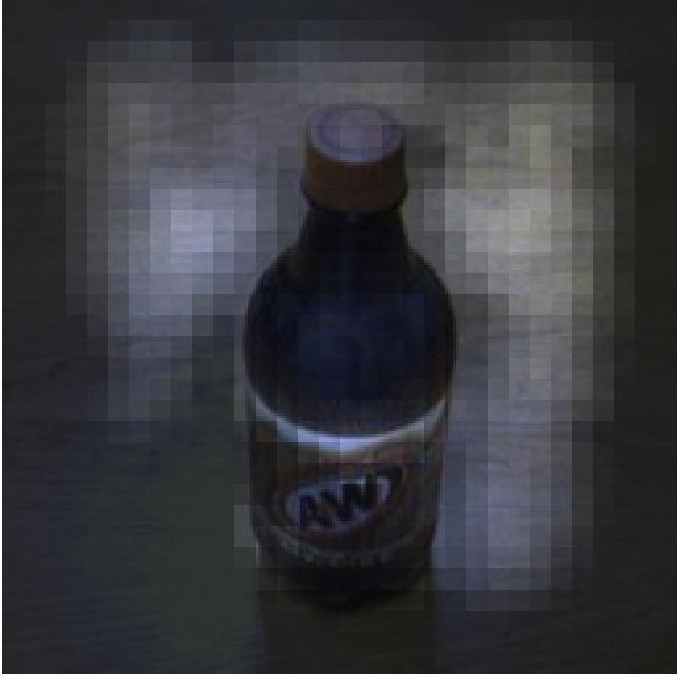}\\ 
\hline\hline
A & \includegraphics[trim=0 -10 0 -10,width=0.09\linewidth]{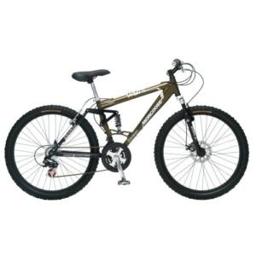} 
& \includegraphics[trim=0 -10 0 -10,width=0.09\linewidth]{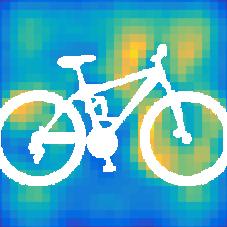} 
& \includegraphics[trim=0 -10 0 -10,width=0.09\linewidth]{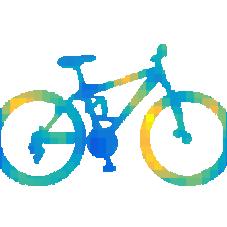} &
A & \includegraphics[trim=0 -10 0 -10,width=0.09\linewidth]{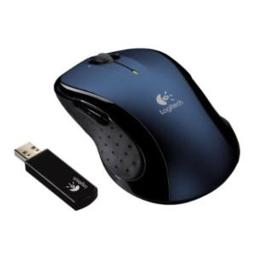} 
& \includegraphics[trim=0 -10 0 -10,width=0.09\linewidth]{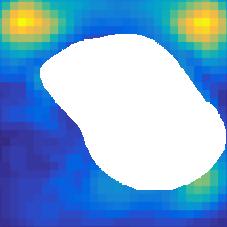} 
& \includegraphics[trim=0 -10 0 -10,width=0.09\linewidth]{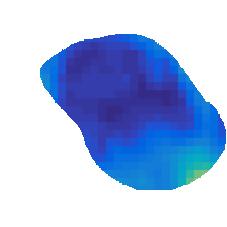} &
D & \includegraphics[trim=0 -10 0 -10,width=0.09\linewidth]{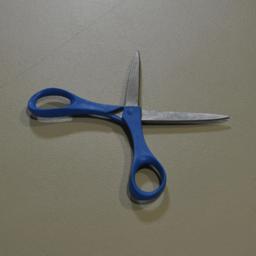} 
& \includegraphics[trim=0 -10 0 -10,width=0.09\linewidth]{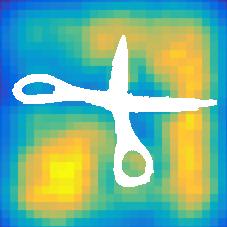} 
& \includegraphics[trim=0 -10 0 -10,width=0.09\linewidth]{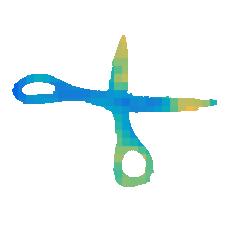} \\ \hline
D & \includegraphics[trim=0 -10 0 -10,width=0.09\linewidth]{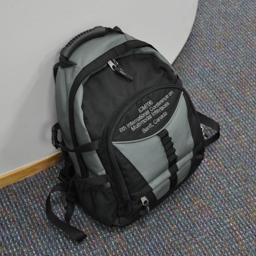} 
& \includegraphics[trim=0 -10 0 -10,width=0.09\linewidth]{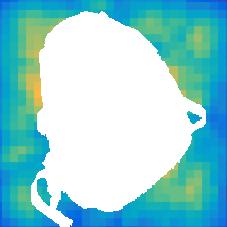} 
& \includegraphics[trim=0 -10 0 -10,width=0.09\linewidth]{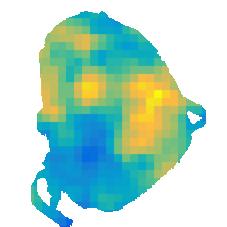} &
W & \includegraphics[trim=0 -10 0 -10,width=0.09\linewidth]{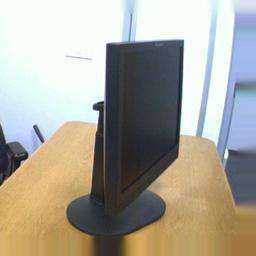} 
& \includegraphics[trim=0 -10 0 -10,width=0.09\linewidth]{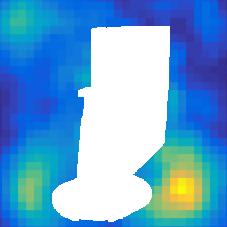} 
& \includegraphics[trim=0 -10 0 -10,width=0.09\linewidth]{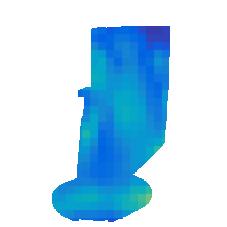} &
W & \includegraphics[trim=0 -10 0 -10,width=0.09\linewidth]{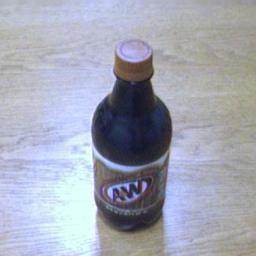} 
& \includegraphics[trim=0 -10 0 -10,width=0.09\linewidth]{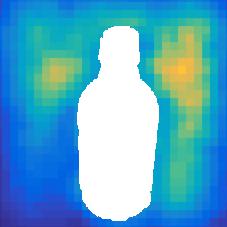} 
& \includegraphics[trim=0 -10 0 -10,width=0.09\linewidth]{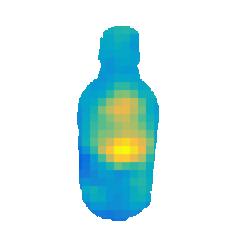}\\ \hline
\end{tabular}
\caption{\textbf{Top}: each table element shows the original image (left), the domainness map (center) and the 
map-image overlap (right - this uses gray tones to occlude low domainness regions. Column titles indicate 
the domain-pairs. \textbf{Bottom}: original image (left), the background part of domainness map (center) 
and the foreground part of the map (right) when using the segmentation masks. Best viewed in color.
}
\label{fig:officem}\vspace{-5mm}
\end{figure}

To have a better understanding of the information captured 
by the domainness maps we analyze here how the domainness distribution in each image relates with 
image foreground and background areas.  

We use the standard Office-31 dataset \cite{Saenko:2010} which collects images from three distinct domains, Amazon (A), Dslr (D) and Webcam (W). 
The first contains images dowloaded from online merchants and mainly present white background, while the second and 
the third are acquired respectively with a high resolution DSL camera and with a low resolution webcam in real settings 
with background and lighting variations. The 31 categories in the dataset consist of objects commonly encountered in 
office settings, such as keyboards, scissors, monitors and telephones. We define train/test split of respectively 
3000/612, 3000/315 and 1000/293 images for the A-W, A-D and W-D pairs and we follow the procedure described in the 
section \ref{sec:strategy} to generate a map for each test image. Some of the obtained maps are shown in Figure 
\ref{fig:officem} - top.

By using object image masks obtained by manual segmentation we evaluated the average domainness value inside and 
outside the objects. Specifically, we focus on the central $227\times227$ area of the image to avoid artifacts that 
might be due to the CNN architecture used. Our evaluation reveals that for A-D and D-W pairs the average 
domainness value is actually higher inside the objects (respectively 0.48 and 0.44) than outside (0.43, 0.41). 
This indicates that most of the domain specific information tend to appear within the foreground rather than 
in the background part of the image. On the other way round, for A-W the background is the main responsible 
of domain shift, with an average domainness value of 0.24 against 0.27 obtained for the object area 
(see Figure \ref{fig:officem} - bottom).

%% file: experiments.tex
\begin{figure}[t]
\begin{center}
\includegraphics[width=0.85\linewidth]{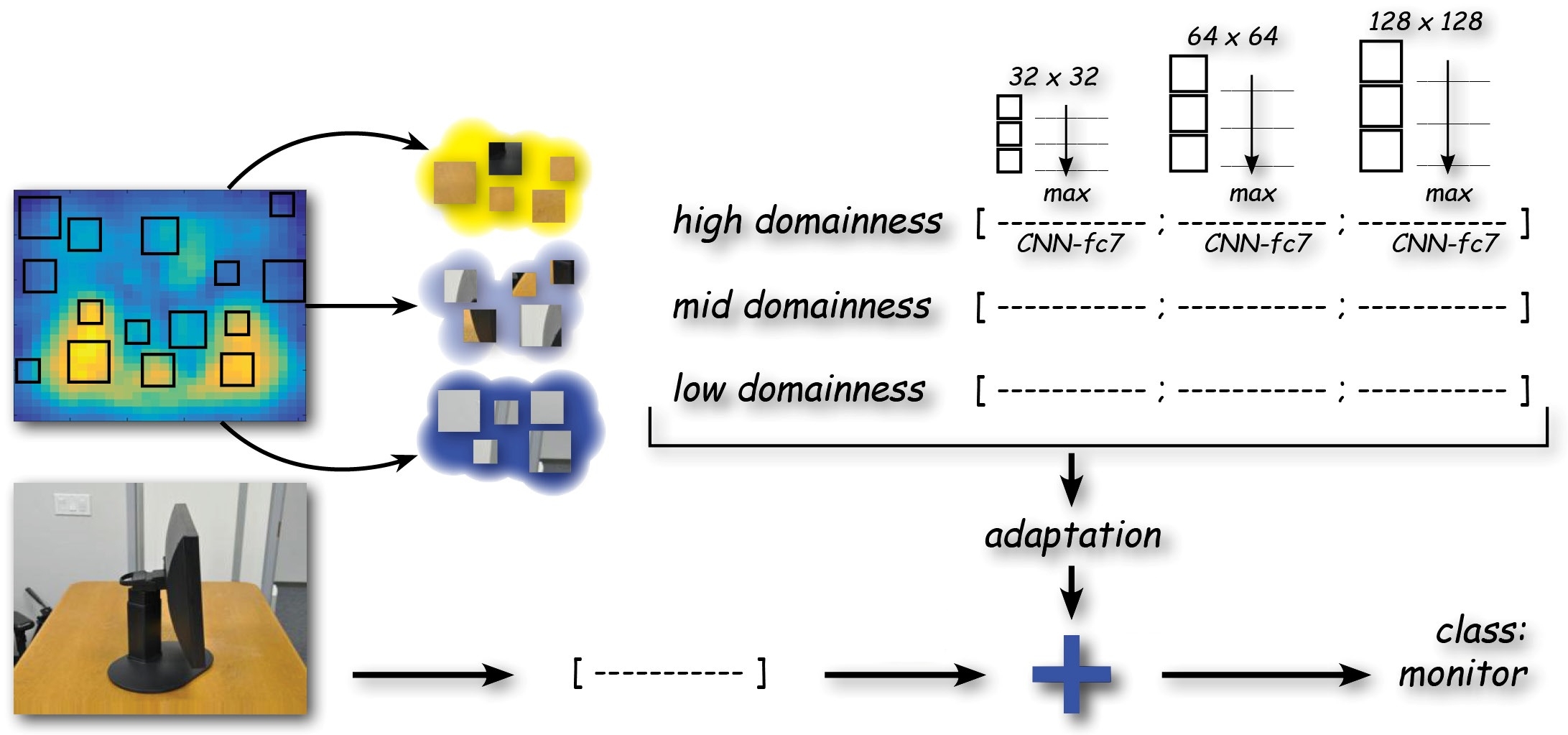} 
\end{center}\vspace{-5mm}
\caption{
Patches of three different dimensions are cropped from an image and 
organized according to their domainness score. CNN-fc7 features are extracted from the patches, 
max pooled and stacked. Classifiers are trained and tested using the obtained representation and 
the whole image. Confidence scores are combined for the final prediction.} \vspace{-4mm}
\label{fig:domainnesslevels}
\end{figure}

\begin{table}[tb]\tiny
\caption{Accuracy on Office-31, usupervised setting - full protocol.}\label{tab:DLexpers}
\vspace{-5mm}
\begin{center}
\begin{tabular}{l c c c c c c |c}
\hline
& ~~~A \vs W~~~ & ~~~A \vs D~~~& ~~~W \vs A~~~ 	& ~~~W \vs D~~~ 	& ~~~D \vs A~~~ 	& ~~~D \vs W~~~ 	& ~~~Average ~~~ \\ 
\hline \hline
G    		&	58.1	&	62.0	&	48.6	&	99.4	&	47.9	&	95.8	& 68.6\\
G-FT		&	60.2	&	60.0	&	49.4	&	99.2	&	49.2	&	95.9	& 69.0\\ \hline
\multicolumn{8}{c}{Domainness Levels (DL)} \\\hline
L-L level   & 44.1		&	49.0	& 28.1		&	90.2	& 32.0		& 84.8		& 54.7 \\
L-H level  	& 41.6		&	42.2	& 27.9		&	77.8	& 30.9		& 73.6		& 49.0\\
L-M level   & 48.9		&	51.0	& 29.9		&	90.8	& 34.5  	& 87.4		& 57.1\\\hline
M-M level 	& 53.0		&	52.1	& 33.3		&	94.1	& 35.1		& 88.6		& 59.4\\
M-L level  	& 45.3		&	50.0	& 31.8		&	89.4	& 31.4		& 84.0		& 55.3\\
M-H level 	& 47.1		&	47.9	& 29.3		&	88.8	& 31.6		& 83.1	    & 54.6 \\\hline
H-H level	& 46.2		&	44.1	& 28.9		&	90.4	& 31.6		& 81.6		& 53.8 \\
H-L level	& 44.1		&	42.6	& 31.0		&	83.3	& 30.3		& 76.5		& 51.3 \\
H-M level 	& 52.4		&	47.4	& 33.1		&	92.5	& 35.1		& 85.7		& 57.7 \\
\hline 
\multicolumn{8}{c}{Applying Domain Adaptation Machine on Domainness Levels (DAM-DL)} \\\hline
L-L level   &	40.8	&	50.0	&	28.4	&	99.9	&	32.9	&	88.9	& 55.7\\
L-H level   &	42.8	&	44.5	&	28.6    &	82.5	&	32.2	&	81.0	& 51.9\\
L-M level   &	48.3	&	50.6	&	31.5	&	92.8	&	34.5	&	91.7	& 58.3\\  \hline
M-M level   &	51.5	&	51.9	&	35.1	&	96.9	&	38.2	&	92.2	& 60.9\\  
M-L level   &	41.8	&	48.8	&	33.7	&	93.8	&	33.8	&	87.5	& 56.6\\
M-H level   &	47.8	&	48.3	&	31.4	&	93.2	&	35.5	&	87.3	& 57.3\\  \hline
H-H level   &	42.7	&	45.8	&	31.2	&	93.2	&	34.3	&	84.3	& 55.2\\
H-L level   &	40.2	&	42.7	&	32.9	&	85.5	&	31.8	&	77.8	& 51.8\\
H-M level   &	47.8	&	50.6	&	34.9	&	94.3	&	36.4	&	87.9	& 58.7\\
\hline 
\multicolumn{8}{c}{Combining Domainness Levels \& Whole Image Classification} \\\hline
G + DL  			&70.6$\pm$0.9	&74.9$\pm$1.1	&53.5$\pm$0.3	&\textbf{100.0}$\pm$0.1	&54.5$\pm$0.5	&98.3$\pm$0.1 &	75.3\\
G + DAM-DL  			&70.6$\pm$1.3	&\textbf{76.9}$\pm$0.4	&54.5$\pm$0.2	&\textbf{100.0}$\pm$0.1	&\textbf{56.6}$\pm$0.5	&\textbf{99.5}$\pm$0.1 &	\textbf{76.3}\\
G-FT  + DL  	&\textbf{71.5}$\pm$0.6	&74.8$\pm$1.2	&54.0$\pm$0.1	&\textbf{100.0}$\pm$0.1	&55.8$\pm$0.8	&97.9$\pm$0.3 &	75.7\\
G-FT  + DAM-DL &71.3$\pm$1.1	&75.3$\pm$1.0	&\textbf{55.4}$\pm$0.3	&\textbf{100.0}$\pm$0.1	&55.2$\pm$0.7	&98.9$\pm$0.3 &	\textbf{76.3}\\ \hline
DDC\cite{Tzeng_arxiv_2015} & 61.8$\pm$0.4 & 64.4$\pm$0.3 & 52.2$\pm$0.4 &98.5$\pm$0.4 &52.1$\pm$0.8 &95.0$\pm$0.5 & 70.6\\
DAN \cite{long_icml_2015}  & 68.5$\pm$0.4 &	67.0$\pm$0.4 & 53.1$\pm$0.3 &99.0$\pm$0.2 &54.0$\pm$0.4 &96.0$\pm$0.3 &	72.9\\
\hline
\end{tabular}
\end{center}\vspace{-5mm}
\end{table}

Finally we use the domainness maps to guide domain adaptation (see Figure \ref{fig:domainnesslevels}). 
Information from the images at three domainness levels (DL) are collected through local feature extraction. 
We start by sampling 100 patches of sizes $32\times32$ randomly from each image and associating to each one its average 
domainness. We then treat the patches as samples of the domainness distribution and identify its 33th and 66th percentiles. 
This allows to divide the patches into three groups with low- (L), mid- (M) and high-domainness (H). 
We follow \cite{Gong_ECCV_2014} and used the CNN Caffe implementation pre-traiened on Imagenet to collect fc7 features for each patch. 
Maximum pooling (element-wise maximum) is applied separately over the patches collected at different domainness levels.
The procedure is repeated separately over other two image scales using patches of dimension $64\times64$ and $128\times128$. 
As a result, each image is represented by three feature vectors, each 
with 12288 elements, obtained by stacking the max pooled features at scale 32, 64 and 128.
Besides these per-domainness-level descriptors we extracted fc7 features on the whole image as global representation  
both with (G-FT) and without (G) fine tuning on the source.

\vspace{1mm} \noindent
\emph{How good are the descriptors?}
We perform object classification with linear SVM on all the domain pairs of the Office-31 dataset when using each DL descriptor to represent the images. 
We consider all the source labeled data as training samples and all the unlabeled target images define our test set (full protocol). 
The results are reported in the top part of Table \ref{tab:DLexpers}, together with the performance obtained using the whole image representation.
The obtained classification accuracies indicate M as the most informative level. 
Although by construction L is the level which capture the most domain-generic cues, we speculate that M works best 
at balancing domain-generic and object-specific information. 

\vspace{1mm} \noindent
\emph{Can we adapt across DLs?}
We want to check if domain adaptation can help to reduce the discrepancy across domainness levels. We use the Domain Adaptation 
Machine (DAM) originally introduced in \cite{DuanICML2009}. 
The results in the central part of Table \ref{tab:DLexpers} show an average accuracy improvement which ranges in 
0.5-3\% with respect to the previous results, confirming that adaptive techniques are still beneficial on the 
defined domainness level representation.  

\vspace{1mm} \noindent
\emph{Are DLs complementary to the whole image?}
The classification performance that we get by using the DL representation is lower than what obtained with 
the full image. Still, we believe that different domainness level provide complementary knowledge that is useful to 
solve domain adaptation. To test this hypothesis we integrate the per-class confidence score provided by the classifiers 
trained over DLs with that obtained when training on the whole image. Let's indicate with $j=1\ldots 9$ the different DL 
pairs and with $c=1\ldots C$ the object classes. Once we have all the margins $D_c^j$ obtained by separate-level classification 
and the margin $D_c^G$ obtained from the whole image we perform the final prediction with 
$\small{c^* = argmax_{c}~\{\frac{1}{9}\sum_{j=1}^9D_c^j +  D_c^G \}~}$.
The obtained results (Table \ref{tab:DLexpers} -- bottom part) compare favorably against the current state of the art methods \cite{Tzeng_arxiv_2015,long_icml_2015} 
based on CNN-architecture created on purpose to overcome visual domain shift.

%% file: conclusion.tex
The goal of this paper is to identify the spatial roots of visual domain shift. To this end
we learned domainnes maps from source and target data which are able to localize the  
image parts more or less responsible for the domain shift. 
We proved experimentally that generating features from image regions with different degrees 
of domainnes and feeding them to a domain adaptation algorithm leads to a significant boost 
in performance. Moreover, in combination with whole image features, they allow to obtain
state of the art results on the Office dataset.

%% file: main.bbl
\begin{thebibliography}{10}

\bibitem{Saenko:2010}
Saenko, K., Kulis, B., Fritz, M., Darrell, T.:
\newblock Adapting visual category models to new domains.
\newblock In: European Conference on Computer Vision - ECCV. (2010)

\bibitem{GongSSG12}
Gong, B., Shi, Y., Sha, F., Grauman, K.:
\newblock Geodesic flow kernel for unsupervised domain adaptation.
\newblock In: IEEE Conference on Computer Vision and Pattern Recognition -
  CVPR. (2012)

\bibitem{long_icml_2015}
Long, M., Cao, Y., Wang, J., Jordan, M.:
\newblock Learning transferable features with deep adaptation networks.
\newblock In: International Conference on Machine Learning - ICML. (2015)

\bibitem{landmarks}
Gong, B., Grauman, K., Sha, F.:
\newblock Connecting the dots with landmarks: Discriminatively learning
  domain-invariant features for unsupervised domain adaptation.
\newblock In: International Conference on Machine Learning - ICML. (2013)

\bibitem{Lim_nips11}
Lim, J.J., Salakhutdinov, R., Torralba, A.:
\newblock Transfer learning by borrowing examples for multiclass object
  detection.
\newblock In: Neural Information Processing Systems - NIPS. (2011)

\bibitem{Hoffman_ECCV2012}
Hoffman, J., Kulis, B., Darrell, T., Saenko, K.:
\newblock Discovering latent domains for multisource domain adaptation.
\newblock In: European Conference on Computer Vision - ECCV. (2012)

\bibitem{reshape}
Gong, B., Grauman, K., Sha, F.:
\newblock Reshaping visual datasets for domain adaptation.
\newblock In: Advances in Neural Information Processing Systems - NIPS. (2013)

\bibitem{DuanICML2009}
Duan, L., Tsang, I.W., Xu, D., Chua, T.S.:
\newblock Domain adaptation from multiple sources via auxiliary classifiers.
\newblock In: International Conference on Machine Learning - ICML. (2009)

\bibitem{DuanCVPR2009}
Duan, L., Tsang, I.W., Xu, D., Maybank, S.J.:
\newblock Domain transfer svm for video concept detection.
\newblock In: IEEE International Conference on Computer Vision and Pattern
  Recognition - CVPR. (2009)

\bibitem{Gretton_MMD_2012}
Gretton, A., Borgwardt, K.M., Rasch, M.J., Sch\"{o}lkopf, B., Smola, A.:
\newblock A kernel two-sample test.
\newblock J. Mach. Learn. Res. \textbf{13}(1) (mar 2012)  723--773

\bibitem{Fernando2013b}
Fernando, B., Habrard, A., Sebban, M., Tuytelaars, T.:
\newblock Unsupervised visual domain adaptation using subspace alignment.
\newblock In: International Conference in Computer Vision - ICCV. (2013)

\bibitem{Hoffman_ICLR2013}
Hoffman, J., Rodner, E., Donahue, J., Saenko, K., Darrell, T.:
\newblock Efficient learning of domain-invariant image representations.
\newblock In: International Conference on Learning Representations - ICLR.
  (2013)

\bibitem{Fernando_Tommasi_Tuytelaars_PRL2015}
Fernando, B., Tommasi, T., Tuytelaars, T.:
\newblock Joint cross-domain classification and subspace learning for
  unsupervised adaptation.
\newblock Pattern Recognition Letters \textbf{65} (nov 2015)  60--66

\bibitem{Tzeng_ICCV_2015}
Tzeng, E., Hoffman, J., Darrell, T., Saenko, K.:
\newblock Simultaneous deep transfer across domains and tasks.
\newblock In: International Conference in Computer Vision - ICCV. (2015)

\bibitem{Tzeng_arxiv_2015}
Tzeng, E., Hoffman, J., Zhang, N., Saenko, K., Darrell, T.:
\newblock Deep domain confusion: Maximizing for domain invariance.
\newblock {arXiv preprint abs/1412.3474} (2014)

\bibitem{Tommasi_2013_ICCV}
Tommasi, T., Caputo, B.:
\newblock Frustratingly easy nbnn domain adaptation.
\newblock In: International Conference on Computer Vision - ICCV. (2013)

\bibitem{Kuzborskij_CVPR_2016}
Kuzborskij, I., Carlucci, F.M., Caputo, B.:
\newblock When na{\"{\i}}ve bayes nearest neighbours meet convolutional neural
  networks.
\newblock In: IEEE Conference on Computer Vision and Pattern Recognition -
  CVPR. (2016)

\bibitem{Zeiler_ECCV_2014}
Zeiler, M.D., Fergus, R.:
\newblock Visualizing and understanding convolutional networks.
\newblock In: European Conference on Computer Vision - ECCV. (2014)

\bibitem{girshick2014rcnn}
Girshick, R., Donahue, J., Darrell, T., Malik, J.:
\newblock Rich feature hierarchies for accurate object detection and semantic
  segmentation.
\newblock In: IEEE Conference on Computer Vision and Pattern Recognition -
  CVPR. (2014)

\bibitem{Simonyan_ICLR_2014}
K.~Simonyan, A.~Vedaldi, A.Z.:
\newblock Deep inside convolutional networks: Visualising image classification
  models and saliency maps.
\newblock In: ICLR Workshop. (2014)

\bibitem{Zhou_ICLR_2015}
Zhou, B., Khosla, A., Lapedriza, A., Oliva, A., Torralba, A.:
\newblock Object detectors emerge in deep scene cnns.
\newblock In: International Conference on Learning Representations - ICLR.
  (2015)

\bibitem{jia2014caffe}
Jia, Y., Shelhamer, E., Donahue, J., Karayev, S., Long, J., Girshick, R.,
  Guadarrama, S., Darrell, T.:
\newblock Caffe: Convolutional architecture for fast feature embedding.
\newblock arXiv preprint abs/1408.5093 (2014)

\bibitem{Gong_ECCV_2014}
Yunchao~Gong, Liwei~Wang, R.G., Lazebnik, S.:
\newblock Multi-scale orderless pooling of deep convolutional activation
  features.
\newblock In: European Conference on Computer Vision - ECCV. (2014)

\end{thebibliography}
